# New Efficient Visual OILU Markers

Youssef Chahir[1] , Messaoud Mostefai[2], Hamza Saidat[2]

[1]Normandie Univ, UNICAEN, ENSICAEN, CNRS, GREYC, 14000 Caen, France
[2]MSE Laboratory, University of Bordj Bou Arreridj, 34000, Algeria

youssef.chahir@unicaen.fr , m.mostefai@univ-bba.dz , hamzasaidat34@gmail.com

**Abstract**
Basic patterns are the source of a wide range of more or less complex geometric structures. We will exploit such patterns to develop new efficient visual markers. Besides being projective invariants, the proposed markers allow producing rich panel of unique identifiers, highly required for resource-intensive navigation and augmented reality applications. The spiral topology of our markers permits the validation of an accurate identification scheme, which is based on level set methods. The robustness of the markers against acquisition and geometric distortions is validated by extensive experimental tests.

## 1. Introduction

Fiducial Markers are widely used in applications where accurate localization within a workspace is required. This is the case, for example, with Unmanned Aerial Vehicles [1] and Augmented Reality applications [2]. Visual markers are not limited to these two fields, but are widely used in different areas such as robotics [3], surgery [4], and human machine monitoring [5]. All of these applications are based on specific patterns, which embed unique real time identifiers. Figure 1, presents the different types of markers for which several dedicated and tutorial articles have been published. Most of these markers belong to the square markers family (Figure1.a) where 2D binary matrices are used to represent various patterns [6]. Since these are dot-matrix markers, any acquisition (lighting, noise etc.) or geometrical distortions will seriously affect the recognition performances and induce inter-marker confusion. A second group of circular markers are developed [7]. The last (Figure1.b), allow precise localization and are less sensitive to noise. Unfortunately, their performances are achieved at the expense of their computation complexity and limited information-coding capacity. Another type of doted [8] and line based [9] projective invariant markers are developed (Figures 1.c, d). Even if they allow rapid and accurate detection, the last generate limited panel of unique identifiers.

Recently a new numbering system called OILU System has been developed, and a dedicated patent was lately filled [10]. It associates basic patterns to decimal digits and allows producing numbers with highly distinguishable patterns. The objective of this short paper is to highlight the effectiveness of such system in the develop-ment of efficient visual markers, less computational and robust to the well-known distortions.

The remainder of this paper is structured as follows. Section 2 sets out the OILU numbering system basics. Section 3 presents the development of visual OILU markers and shows their characteristics. Sections 4 describes the markers detection and identification approach. The obtained test results are displayed in section 5. The conclusion and future work are reported in section 6.

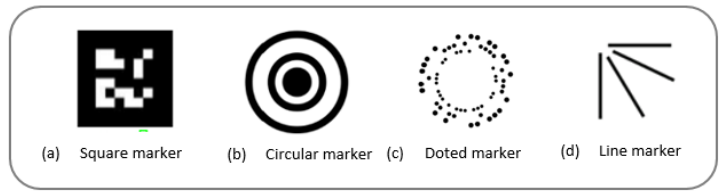

**Figure 1**. Example of Visual Markers

## 2. The OILU Numbering System Basics

The proposed approach uses four basic patterns for the generation of a new decimal numbering system. The latter is composed of ten symbols, which are generated as follows: the first four basic symbols {O, I, L and U} are respectively affected to digits zero, one, two and three (Figure 2.a). Whereas, the remaining symbols are then generated by successive rotations (quarter counterclockwise) of the two symbols L and U (Figure 2b).

The main interesting thing with this symbolic is that, it allows superimposing symbols in pyramidal form without losing the value of the constructed numbers. As an illustration example, a pyramidal composition of the decimal number 4670 is shown in Figure 3a. The lecture sense is from the outside to the inside. Unlike the classical numeral decimal symbolic, the OILU symbolic allows to see a number as an object with its four facets. Indeed, we can extract from each point of view a different number value. Thus, a group of related numbers is formed: 4670 – 2450 – 8230 – 6890. Note that any facet's value allows deducing the other facets values. Multi facets numbers have interesting applications in the large field of data processing, but in our case, they are mainly exploited for the design of new efficient square OILU markers.

## 3. Square OILU Markers Design

OILU symbolic allows producing visual makers composed of a group of superimposed OILU symbols. The marker characteristics (embedded OILU number, size, color and symbol's lines thickness) are fixed according to the application requirements (Robustness to motion blur, distance from camera, etc.). For a better visibility and real time performances, symbols lines are designed according to the surrounding space characteristics: black lines on white background or white lines on black background (Figure 3.b). Figures 3.c, 3.d and 3.e present examples of neighboring OILU markers as well as their associated codification table.

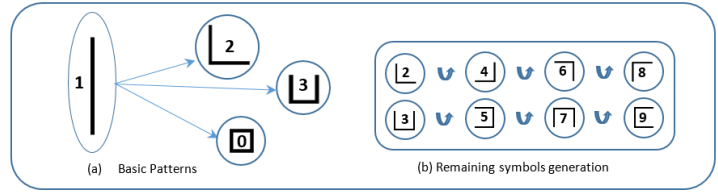

**Figure 2.** OILU Symbols Generation

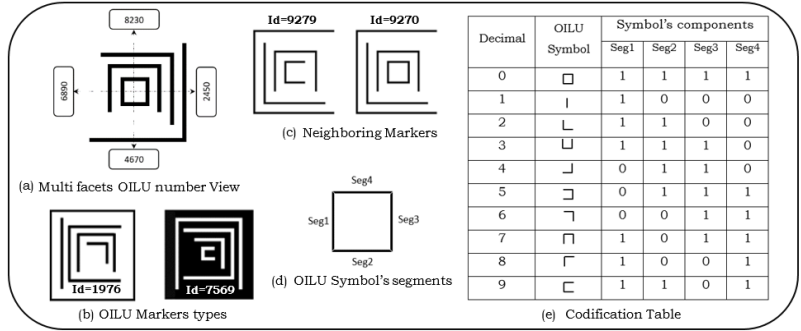

**Figure 3.** OILU Markers Presentation and Codification

## 4. OILU Markers Detection and Identification

Acquired images are converted to gray scale and then binarized. A cleanup operation is applied to remove all connected components that have less than a certain number of connected pixels. The marker boundaries are then detected and the region of interest is cropped. The following step consists on generating embedded code from groups of parallel-line segments. Such task can be considered as a geometrical problem that can be solved with space filling and distance maps methods [11]. For a first approach, we exploit the OILU markers topology which is similar to Fermat's spirals [12] to generate the corresponding distance map using level-set methods [13,14]. In this paper, we used a variant of Fermat spirals as a 2D filling model (see Figure 4). The level-set method offers a more straightforward solution for offset geometric structures to generate distance maps. In this method, the quadrilateral border ($\Omega$) is represented by an implicit function $\Phi$, which has the following properties:

$$\Phi(p) = \begin{cases} -d(p) = W & if\ p\ \in \Omega \\ 0 & if\ p\ \in \partial\Omega \\ d(p) = -W & if\ p\ \notin \Omega \end{cases} \quad \text{where d(.) is a distance function.}$$

Once the marker distance map generated, an inward propagation is performed to label the pixels inside the quadrilateral, according to the level set or turns (see Figure 5). To extract the embedded code, we do the following:

1. Divide the main quadrilateral into 4 triangles, obtained from the diagonals. Their analysis allows knowing the presence of OILU segments.

2.  For each triangle we do the following:
    -   Check whether a labeled point lies inside a triangle or not;
    -   Count labeled points inside a triangle. If the number of points exceeds a fixed threshold, a segment is considered to be present. It will be represented by '1', otherwise '0'.

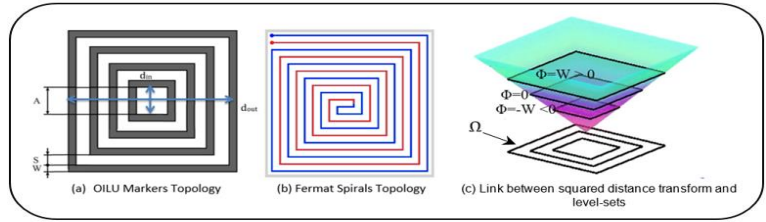

**Figure 4**. OILU Markers Topology and corresponding distance Map Generation

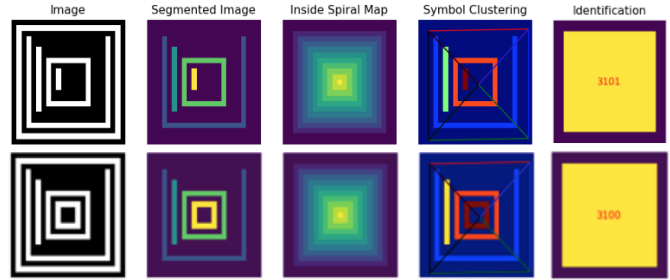

**Figure 5.** OILU Markers identification

## 5.  Preliminary Experimental Tests

First tests were carried on synthetic markers, which were gradually degraded using several well-known distortions (additive noise, blur and radial distortion). Obtained robustness tests are presented in figure 6.a. As can be seen, markers identification scheme demonstrated satisfactory results even with high levels of distortions. This is mainly due to the fact that segments labeling is based on a pre-established distance map.

The second series of tests were performed in real environment using two different size markers M1 (12x12cm) and M2 (6x6cm). Initially, we evaluated the impact of the marker to camera distance on the performances of marker detection and recognition. The camera was positioned at different distances from approximately 0.5 m to 6m. The distance range at which the marker was reliably detectable and identifiable was between 0.5m and 3m for the marker M2, and extended to 5m for the marker M1(Figure 6.b).

The third test concerns the robustness to viewing angle (Figure 6.c). OILU markers were acquired with variable angles of view, ranging from approximately 5° to 30°. Marker identification was successful only for angles superior than 10°. Bellow this angle, the segments were very close, which made identification very difficult. Finally, we tested the algorithm in difficult lighting conditions. As can be seen (Figure 6.d), below a certain lighting level (25 Lux), identification became also difficult.

The experiments have been performed on a typical Mac Book, equipped with a 2,9 GHz Intel Core i5 processor and 8 Go of RAM. Detection and identification performances were evaluated by computing the processing time for marker detection and identification. The latter was around 40 ms. This result showed that even if Level Set methods were efficient, they remain computationally expensive. We are currently working on a faster method, which allows real-time identification of the targeted marker. At this stage, we continue to improve the performances of this method, especially against occlusion. A demonstration of the initial version is available at [15].

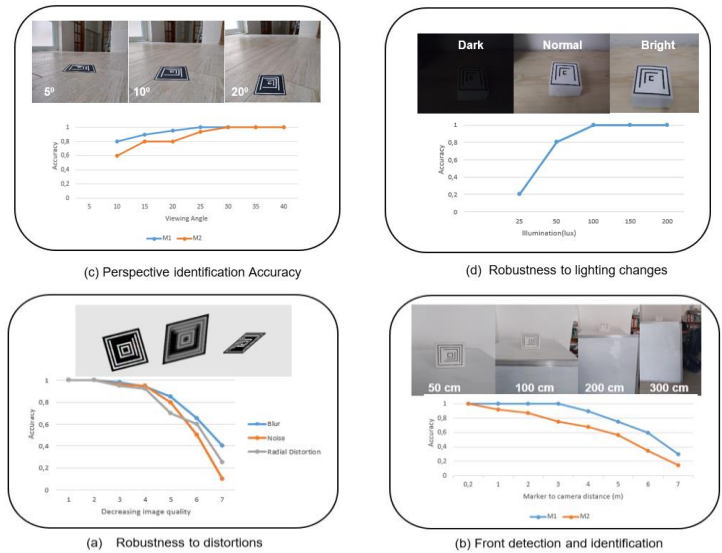

**Figure 6.** Robustness Tests

## 6. Conclusion and Future Work

We have presented the development of a new visual marker based on the OILU numbering system. Presented preliminary tests aim at validating the OILU marker functionality. However, deep comparative tests will be required to accurately evaluate the OILU marker's performances against leading state of the art markers. Additionally, software improvements are necessary to allow robust real-time identification and pose estimation.

It is important to note that the proposed OILU system allows generating multi types (square, circular and doted) markers with identical coding capacities. Indeed, replacing segments by arcs or dots will allow producing circular or doted OILU markers (Figure 7). These interesting options will be addressed in a future work.

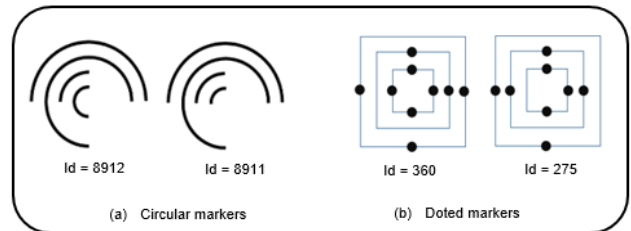

**Figure 7.** Examples of circular and doted OILU Markers